\documentclass[11pt]{article}

\usepackage[margin=1in]{geometry}

\usepackage{amsmath}
\usepackage{amssymb}
\usepackage{amsfonts}
\usepackage{mathtools}

\usepackage{graphicx}
\usepackage{tikz}
\usepackage{booktabs}
\usepackage{array}
\usepackage{caption}
\usepackage{subcaption}
\usepackage{float}
\usetikzlibrary{
    arrows.meta,
    positioning,
    fit,
    backgrounds,
    shapes.geometric,
    decorations.pathreplacing,
    calc
}

\usepackage{algorithm}
\usepackage{algpseudocode}

\usepackage[numbers,sort&compress]{natbib}

\usepackage{xcolor}
\usepackage{hyperref}
\usepackage[nameinlink,capitalize]{cleveref}

\hypersetup{
    colorlinks=true,
    linkcolor=blue,
    citecolor=blue,
    urlcolor=blue
}

\newcommand{\method}{GRAPE}
\newcommand{\pgdat}{AT}
\newcommand{\iseat}{ISEAT}
\newcommand{\loss}{\mathcal{L}}
\newcommand{\data}{\mathcal{D}}

\title{
\method{}: Guided Parameter-Space Evolution\\
for Compact Adversarial Robustness
}

\author{
Zhiyuan Ye\textsuperscript{1},
Xiangyu Zhou\textsuperscript{2},
Ji Qi\textsuperscript{2,*},
Hao Zhang\textsuperscript{1,*},
and Yi Zhou\textsuperscript{2,*}
\\[0.35em]
\textsuperscript{1}University of Science and Technology of China
\\
\textsuperscript{2}China Mobile (Suzhou) Software Technology Co., Ltd.
\\[0.35em]
\texttt{zhiyuan.ye@mail.ustc.edu.cn}
\\
\texttt{\{zhouxiangyu,qiji\}@cms.chinamobile.com}
\\
\texttt{\{hzhangzy,yi\_zhou\}@ustc.edu.cn}
\\[0.2em]
\textsuperscript{*}Corresponding authors.
}

\date{}

\begin{document}

\maketitle

\begin{abstract}
Adversarial Training (AT) is a mainstream approach for improving neural network robustness, but most AT methods assume that the optimizable parameter space remains fixed throughout training. Although larger model capacity often improves robust accuracy, the robustness supported by a model should not be determined solely by its static parameter count before training. Inspired by the high-level principle of developmental plasticity, where representational pathways are progressively formed, stabilized, and reorganized, this paper studies whether the order in which the parameter space is exposed during training can affect the final robust solution, even when the final architecture is unchanged.

We propose \method{}, namely Guided Parameter-Space Evolution, a training framework for compact adversarial robustness. \method{} combines parameter-space stabilization with progressive hidden expansion: the former stabilizes local robust optimization, while the latter gradually releases new optimizable dimensions during training. Unlike fixed-structure AT, which exposes the full parameter space from the beginning, \method{} treats robust model learning as a process of progressive parameter-space exposure and evolution. It further uses an adversarial spectral utilization score to allocate newly released capacity to high-pressure modules, improving the efficiency of parameter-space expansion.

Under the standard $\ell_\infty$ threat model on CIFAR-10, using fixed-structure ResNet-18 AT as a controlled reference, \method{} improves PGD-20 robust accuracy from 51.70\% to 56.94\% at a nearly matched computation budget with a FLOPs ratio of $1.009\times$, corresponding to a relative improvement of approximately 10.1\%, while reducing the parameter count by approximately 21.4\%. Moreover, a sequential grow variant that preserves the final ResNet-18 architecture reaches 56.52\% PGD-20 robust accuracy, indicating that the gain is not solely due to final architecture differences, but is also related to the progressive exposure path of the parameter space. These results suggest that guided parameter-space evolution can form a more compact and more robust parameter-space configuration under matched computation.
\end{abstract}

\section{Introduction}

Deep neural networks have achieved remarkable success in image classification and related tasks, but they remain highly sensitive to small adversarial perturbations. Given an input sample $x$, label $y$, model $f_\theta$, and perturbation budget $\epsilon$, standard Adversarial Training (AT) is commonly formulated as the following min-max optimization problem:
\begin{equation}
\min_{\theta}
\mathbb{E}_{(x,y)\sim\data}
\left[
\max_{\|\delta\|_\infty \leq \epsilon}
\loss(f_\theta(x+\delta), y)
\right].
\label{eq:pgdat}
\end{equation}

In this paper, we use \pgdat{} to denote standard adversarial training where the inner maximization is approximated by multi-step PGD. This type of method is one of the most widely used paradigms in empirical robustness research \citep{madry2018towards}. Subsequent studies have further improved adversarial training from the perspectives of the robustness-accuracy trade-off \citep{zhang2019trades}, robust overfitting \citep{rice2020overfitting}, loss landscape smoothing \citep{wu2020awp}, data augmentation, and regularization.

Among these directions, parameter-space stabilization is an important approach for improving adversarial training. AWP introduces adversarial weight perturbations in the weight space, encouraging the model to converge to flatter robust solutions \citep{wu2020awp}. \iseat{} further introduces adaptive smoothing according to instance-wise vulnerability, thereby stabilizing the local optimization process around highly vulnerable samples \citep{li2023iseat}. These methods indicate that adversarial training depends not only on perturbation construction in the input space, but is also closely related to local stability in the parameter space. However, they usually share an implicit assumption: the model architecture and the optimizable parameter space remain fixed throughout training.

It is widely observed in adversarial robustness that larger model capacity often helps improve robust accuracy. However, the robustness that a model can support should not be determined solely by the static total number of parameters specified before training. Prior work also shows that robustness does not simply increase monotonically with model capacity. Huang et al. systematically studied the effects of width and depth configurations in adversarially trained DNNs, and observed that more parameters do not necessarily improve robustness; in some cases, reducing the capacity of the last stage can even improve adversarial robustness \citep{huang2021architectural}. This observation suggests that the effective capacity of an adversarially robust model should not be understood merely as the static parameter count, but should also take into account how the parameter space is exposed, stabilized, and expanded during training.

From the perspective of parameter space, fixed-structure AT can be written as:
\begin{equation}
\theta_t \in \Theta,
\qquad
\Theta = \mathrm{constant}.
\end{equation}
That is, the entire training process is performed within the same parameter space $\Theta$. This paper focuses on a different question: even if the final network architecture is the same, can the order in which the parameter space is exposed during training affect the final robust solution? If the model exposes the full parameter space from the beginning of training, the optimizer may prematurely exploit high degrees of freedom and form an inefficient robust fitting path. In contrast, if the optimizable parameter space is progressively exposed and expanded during training, the model may first form relatively stable robust representations in a smaller parameter space, and then gradually absorb new learnable degrees of freedom. In other words, the evolution path of the parameter space itself may become a key factor affecting the efficiency and robustness of adversarial training.

This perspective can also be motivated by a high-level abstraction from brain-inspired intelligence. Learning in biological neural systems does not occur in a static capacity system that is fully unfolded from the beginning. Instead, representational pathways are progressively formed, stabilized, and reorganized through development and experience. Studies on neural plasticity show that development and learning induce continuous structural and functional plasticity in the nervous system \citep{galvan2010neuralplasticity}. Meanwhile, the stability-plasticity dilemma states that a learning system must balance the stability of existing knowledge with the ability to absorb new information; Adaptive Resonance Theory has also long regarded stable learning and continual adaptation as important mechanisms of cognitive systems \citep{carpenter1987massively,grossberg2013adaptive}. This paper does not aim to simulate specific neural mechanisms. Instead, we borrow this high-level organizational principle and abstract adversarially robust model learning as a process of progressive parameter-space exposure, stabilization, and expansion.

Based on this perspective, we propose \method{}, namely Guided Parameter-Space Evolution, a parameter-space evolution training method for compact adversarial robustness. Unlike fixed-structure \pgdat{}, \method{} formulates robust training as optimization in a dynamic parameter space:
\begin{equation}
\Theta_0 \subset \Theta_1 \subset \cdots \subset \Theta_T.
\end{equation}
Specifically, \method{} combines parameter-space stabilization with progressive hidden expansion. The former borrows the local smoothing idea from AWP and \iseat{} to stabilize the local parameter space during adversarial training, while the latter progressively releases new optimizable dimensions during training. Unlike fixed-structure AT, which exposes the full parameter space from the beginning, progressive expansion allows the model to gradually obtain new hidden capacity during training.

On this basis, \method{} further uses an adversarial spectral utilization score to guide newly released capacity toward high-capacity-pressure modules. In other words, \method{} not only considers whether the parameter space should be progressively expanded, but also where the newly released optimizable dimensions should be allocated. In this paper, a stabilized sequential grow variant without score guidance is used as a control for progressive parameter-space exposure, while score-guided grow serves as the guided parameter-space evolution mechanism. The former is used to test whether progressive parameter-space exposure itself is effective under the same stabilization setting, and the latter is used to test whether adversarial spectral utilization can further improve the efficiency of capacity release.

It is important to emphasize that the goal of \method{} is neither simply to enlarge model capacity nor to treat parameter reduction itself as the primary objective. Since hidden expansion in different blocks leads to imbalanced growth in parameter count and FLOPs, this paper uses FLOPs as the primary computation-budget reference and reports parameter count changes to characterize parameter-space compactness. In other words, we ask whether guided parameter-space evolution can form a compact parameter-space configuration with fewer parameters and higher robustness under the same computation budget.

The contributions of this paper are summarized as follows:
\begin{itemize}
    \item We revisit adversarial training from the perspective of parameter-space evolution, and argue that the capability of a robust model should not be explained solely by the static total number of parameters, but should also take into account how the optimizable parameter space is progressively exposed and expanded during training.
    \item We propose \method{}, namely Guided Parameter-Space Evolution, which combines parameter-space stabilization with progressive hidden expansion, enabling adversarial training to incorporate both local parameter-space stabilization and global parameter-space expansion.
    \item We verify the role of progressive parameter-space exposure itself through a sequential grow variant that preserves the final ResNet-18 architecture. This variant reaches 56.52\% PGD-20 robust accuracy under a final architecture equivalent to ResNet-18, indicating that the gain does not come solely from differences in the final architecture, but is related to the parameter-space exposure path during training.
    \item Under the standard $\ell_\infty$ threat model on CIFAR-10, at a nearly matched computation budget with a FLOPs ratio of $1.009\times$, \method{} improves PGD-20 robust accuracy from 51.70\% for fixed-structure ResNet-18 AT to 56.94\%, corresponding to a relative improvement of approximately 10.1\%, while reducing the parameter count by approximately 21.4\%. This result suggests that guided parameter-space evolution can form a more compact and more robust parameter-space configuration under the same computation budget.
\end{itemize}

\section{Related Work}

\subsection{Adversarial Training}

Adversarial training is a central approach for empirically improving the robustness of deep models. Early studies showed that deep neural networks are vulnerable to adversarial examples that are nearly imperceptible to humans \citep{szegedy2014intriguing,goodfellow2015explaining}. Subsequently, \citet{madry2018towards} formulated adversarial robustness as a robust optimization problem and used multi-step PGD to approximate the inner maximization, establishing the widely used \pgdat{} training paradigm in empirical robustness research. TRADES further models the trade-off between robustness and natural accuracy by separating the clean loss from boundary regularization \citep{zhang2019trades}.

Although these methods substantially improve empirical robustness, AT still faces challenges such as training instability, robust overfitting, and high computational cost. \citet{rice2020overfitting} showed that, during adversarial training, the model may continue to improve on training-set adversarial examples while test robust accuracy stagnates or even degrades. Subsequent work has further improved the practical performance of AT through data augmentation, regularization, model averaging, and training recipes \citep{rebuffi2021fixing,pang2021bag}. These studies mainly focus on the objective, data, or optimization recipe, whereas this paper studies a different question: whether the optimizable parameter space during training should remain fixed.

\subsection{Parameter-Space Stabilization for Robust Training}

Beyond input-space perturbations, local stability in the parameter space is also closely related to adversarial robustness. AWP constructs adversarial weight perturbations in the weight space, encouraging the model to converge to flatter robust solutions and thereby mitigating robust overfitting \citep{wu2020awp}. Relatedly, \iseat{} introduces adaptive smoothing based on instance-wise vulnerability, stabilizing the local optimization process around highly vulnerable samples \citep{li2023iseat}.

In this paper, we regard methods such as AWP and \iseat{} as representatives of parameter-space stabilization: they mainly stabilize adversarial training by smoothing or perturbing the weight space. However, these methods typically still optimize within a fixed architecture and a fixed parameter space. GRAPE borrows the idea of local stabilization from these methods, but further introduces progressive hidden expansion, so that adversarial training is no longer restricted to a pre-fixed parameter space but instead progressively exposes and expands optimizable dimensions during training.

\subsection{Model Capacity and Robust Architecture Design}

The relationship between model capacity and adversarial robustness is complex. Empirically, wider or larger models often achieve higher robust accuracy, especially under strong adversarial training and large-scale data augmentation settings, where model scale often becomes a key factor for improving robustness \citep{gowal2020uncovering,bartoldson2024adversarial}. However, capacity is not a one-dimensional concept. Parameter count, FLOPs, width, depth, and channel allocation across different stages can all affect robustness and computational efficiency.

\citet{huang2021architectural} systematically studied the effects of depth, width, and architectural ingredients in adversarially trained DNNs, and observed that more parameters do not necessarily lead to higher robustness; in some cases, reducing the capacity of the last stage can even improve adversarial robustness. This is consistent with the perspective of this paper: robust capability should not be explained solely by the static total number of parameters, but should also account for how the parameter space is organized, exposed, and expanded. Unlike static architecture search or manual width adjustment, GRAPE focuses on parameter-space evolution during training and analyzes whether a more compact and more robust parameter-space configuration can be formed under a matched FLOPs budget.

Adversarial pruning provides another related route toward compact robust models. Pruning-based methods aim to remove parameters, channels, or neurons while preserving or inducing adversarial robustness \citep{sehwag2020hydra,jian2022pruning,piras2024adversarial}. These methods typically start from a full or over-parameterized model and reduce capacity, so their parameter-space trajectory can be viewed as:
\begin{equation}
\Theta_{\mathrm{large}}
\rightarrow
\Theta_{\mathrm{compact}}.
\end{equation}
In contrast, GRAPE follows the opposite direction. It starts from a smaller parameter space and progressively releases new hidden capacity during adversarial training:
\begin{equation}
\Theta_{\mathrm{small}}
\rightarrow
\Theta_{\mathrm{grown}}.
\end{equation}
Both directions can lead to compact robust models, but they impose different optimization paths. Pruning asks how to preserve or induce robustness after capacity is removed from an existing parameter space, whereas GRAPE asks whether gradually exposing capacity during training can lead to a different and more compact robust solution under a matched computation budget. Therefore, GRAPE should not be viewed as a pruning or compression method, but as a training-time parameter-space evolution method whose compactness emerges from the capacity release path.

\subsection{Progressive Growth and Network Morphism}

Dynamically expanding network structures is not a new problem. Net2Net expands a smaller network into a wider or deeper one through function-preserving transformations, thereby accelerating model transfer and continued training \citep{chen2016net2net}. Network Morphism further studies how to change the network architecture while preserving the original function, allowing the model to smoothly transition from an existing solution to a larger architecture \citep{wei2016network}. These works mainly target accelerated training or architecture transfer in standard supervised learning, rather than robustness-computation efficiency in adversarial training.

Another related line of work is neural network growth or neurogenesis. \citet{maile2022when} studied when, where, and how to add new neurons during neural network training, and analyzed the effect of dynamic growth on accuracy and network size. These methods show that expanding the parameter space during training can itself change the optimization trajectory. GRAPE shares the idea of progressive expansion with these works, but studies a different setting: this paper focuses on adversarial training and further examines whether the parameter-space exposure path still affects the final robust solution when the final architecture is the same. Moreover, GRAPE uses an adversarial spectral utilization score to guide newly released capacity toward high-pressure modules, thereby connecting dynamic growth with capacity pressure in adversarially robust training.

\subsection{Spectral Utilization and Effective Dimensionality}

This paper uses an adversarial spectral utilization score to estimate the capacity pressure of different modules. This score is based on the spectral distribution of representations and effective rank. Effective rank was proposed by \citet{roy2007effective} to characterize the effective dimensionality of a matrix through spectral entropy. Compared with simply observing channel count or parameter count, effective rank can reflect the effective dimensions actually used in the representation space.

In representation learning, spectral distributions, covariance structures, and dimensional utilization are also commonly used to analyze representation collapse or redundancy. For example, VICReg uses variance and covariance regularization to avoid collapse in self-supervised representations \citep{bardes2022vicreg}. This paper does not directly use such regularization terms. Instead, we use the spectral utilization of adversarial features as a grow score to estimate the capacity pressure borne by different residual blocks during adversarial training. In this way, GRAPE transforms effective-dimensionality analysis in the representation space into a decision signal for parameter-space expansion.

\subsection{Brain-Inspired Plasticity}

The brain-inspired motivation of this paper comes from developmental plasticity and the stability-plasticity abstraction, rather than from a direct simulation of specific neural mechanisms. Studies on neural plasticity show that development and learning induce continuous structural and functional plasticity in the nervous system \citep{galvan2010neuralplasticity}. Meanwhile, the stability-plasticity dilemma emphasizes that a learning system must balance the stability of existing knowledge with the ability to absorb new information; Adaptive Resonance Theory has also long regarded stable learning and continual adaptation as important mechanisms of cognitive systems \citep{carpenter1987massively,grossberg2013adaptive}.

GRAPE brings this high-level organizational principle into adversarial training. It uses local parameter-space stabilization to preserve existing robust representations. It uses progressive hidden expansion to release new optimizable dimensions. It also uses score-guided allocation to decide which high-pressure modules should receive newly released capacity. This paper does not claim to simulate biological neural systems. Instead, it borrows the organizational principle of ``stabilizing existing representations + progressively releasing plastic capacity'' to provide a parameter-space evolution framework for compact adversarially robust models.

\section{Method}

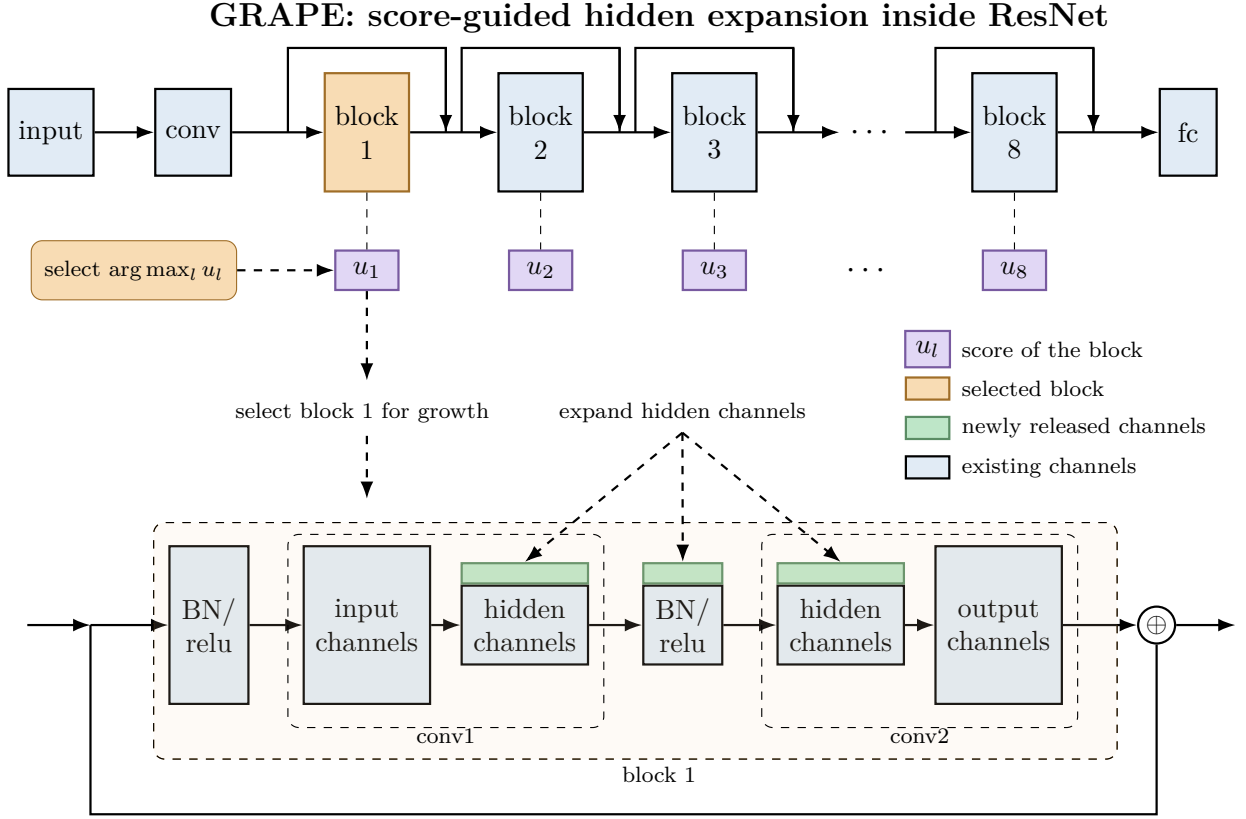
\begin{figure}[t]
\centering
\resizebox{\textwidth}{!}{
\begin{tikzpicture}[
    font=\small,
    >=Latex,
    linegray/.style={draw={rgb,255:red,90; green,90; blue,90}},
    arrow/.style={-{Latex[length=2.2mm]}, thick},
    dashedarrow/.style={-{Latex[length=2.2mm]}, thick, dashed},
    basebox/.style={fill={rgb,255:red,225; green,235; blue,245}},
    selectbox/.style={fill={rgb,255:red,248; green,220; blue,180}, draw={rgb,255:red,160; green,110; blue,40}},
    scorebox/.style={fill={rgb,255:red,225; green,215; blue,245}, draw={rgb,255:red,120; green,90; blue,160}},
    growfill/.style={fill={rgb,255:red,190; green,230; blue,200}, draw={rgb,255:red,90; green,140; blue,100}},
    box/.style={
        rectangle,
        draw=black,
        thick,
        align=center,
        minimum height=0.95cm,
        minimum width=0.75cm
    },
    growbox/.style={
        rectangle,
        draw={rgb,255:red,90; green,140; blue,100},
        thick,
        fill={rgb,255:red,190; green,230; blue,200},
        minimum height=0cm
    },
    note/.style={
        rectangle,
        rounded corners,
        draw=black,
        align=center,
        font=\scriptsize,
        minimum width=2.8cm,
        minimum height=0.75cm
    }
]

\node[font=\bfseries\large] at (1.7, 5.6)
{\method: score-guided hidden expansion inside ResNet};


\node[box, basebox, minimum width=0.90cm, minimum height=1.10cm] (input) at (-6, 4.15) {input};
\node[box, basebox, minimum width=0.72cm, minimum height=1.10cm] (conv)  at (-4.2, 4.15) {conv};

\node[box, selectbox, minimum width=0.90cm, minimum height=1.50cm] (b1) at (-2, 4.15) {block\\1};
\node[box, basebox, minimum width=0.90cm, minimum height=1.50cm] (b2) at (0.2, 4.15) {block\\2};
\node[box, basebox, minimum width=0.90cm, minimum height=1.50cm] (b3) at (2.4, 4.15) {block\\3};

\node[font=\large] (dots1) at (4.4, 4.15) {$\cdots$};

\node[box, basebox, minimum width=0.90cm, minimum height=1.50cm] (b8) at (6.2, 4.15) {block\\8};
\node[box, basebox, minimum width=0.72cm, minimum height=1.10cm] (fc) at (8.4, 4.15) {fc};

\draw[arrow] (input.east) -- (conv.west);
\draw[arrow] (conv.east) -- (b1.west);
\draw[arrow] (b1.east) -- (b2.west);
\draw[arrow] (b2.east) -- (b3.west);
\draw[arrow] (b3.east) -- (dots1.west);
\draw[arrow] (dots1.east) -- (b8.west);
\draw[arrow] (b8.east) -- (fc.west);

\draw[thick]
    (-3,4.15) -- (-3,5.22) -- (-1,5.22) -- (-1,4.15);
\draw[arrow] (-1,5.22) -- (-1,4.15);
\draw[thick]
    (-0.8,4.15) -- (-0.8,5.22) -- (1.2,5.22) -- (1.2,4.15);
\draw[arrow] (1.2,5.22) -- (1.2,4.15);
\draw[thick]
    (1.40,4.15) -- (1.40,5.22) -- (3.40,5.22) -- (3.40,4.15);
\draw[arrow] (3.40,5.22) -- (3.40,4.15);
\draw[thick]
    (5.20,4.15) -- (5.20,5.22) -- (7.20,5.22) -- (7.20,4.15);
\draw[arrow] (7.20,5.22) -- (7.20,4.15);

\node[box, scorebox, minimum width=0.8cm, minimum height=0.5cm] (u1box) at (-2,2.4) {$u_1$};
\node[box, scorebox, minimum width=0.8cm, minimum height=0.5cm] (u2box) at (0.2,2.4) {$u_2$};
\node[box, scorebox, minimum width=0.8cm, minimum height=0.5cm] (u3box) at (2.4,2.4) {$u_3$};
\node[font=\large] at (4.35,2.4) {$\cdots$};
\node[box, scorebox, minimum width=0.8cm, minimum height=0.5cm] (u8box) at (6.2,2.4) {$u_8$};

\draw[dashed] (b1.south) -- (u1box.north);
\draw[dashed] (b2.south) -- (u2box.north);
\draw[dashed] (b3.south) -- (u3box.north);
\draw[dashed] (b8.south) -- (u8box.north);

\node[note, selectbox, minimum width=2.6cm] (selectutil) at (-4.95,2.4)
{select $\arg\max_l u_l$};
\draw[dashedarrow] (selectutil.east) -- (u1box.west);
\draw[dashedarrow] (u1box.south) -- (-2.0,1.0) coordinate (b1select);
\node[font=\scriptsize, anchor=west] (selectgrow) at (-3.8, 0.6) {select block 1 for growth};
\draw[dashedarrow] (-2,0.3) -- ++(0,-0.8);


\coordinate (rin)  at (-5.5, -2.1);
\coordinate (addp) at (8, -2.1);
\coordinate (rout) at (9, -2.1);

\draw[arrow] (-6.3,-2.1) -- (rin);

\node[circle, draw=black, thick, inner sep=1.1pt] (add) at (addp) {$\oplus$};
\draw[arrow] (add.east) -- (rout);

\node[box, basebox, minimum width=1cm, minimum height=2cm] (bn) at (-4, -2.1) {BN/\\relu};
\node[box, basebox, minimum width=1.6cm, minimum height=2cm] (conv1) at (-2, -2.1) {input\\channels};

\node[box, basebox, minimum width=1.6cm, minimum height=1cm] (hidden1) at (0, -2.1) {hidden\\channels};

\node[box, basebox, minimum width=1cm, minimum height=1cm] (bn2) at (2, -2.1) {BN/\\relu};
\node[box, basebox, minimum width=1.6cm, minimum height=1cm] (hidden2) at (4, -2.1) {hidden\\channels};

\node[box, basebox, minimum width=1.6cm, minimum height=2cm] (conv2) at (6, -2.1) {output\\channels};

\node[growbox, anchor=south, minimum width=1.6cm, minimum height=0.20cm] (grow1) at (hidden1.north) {};
\node[growbox, anchor=south, minimum width=1cm, minimum height=0.2cm] (growbn2) at (bn2.north) {};
\node[growbox, anchor=south, minimum width=1.6cm, minimum height=0.24cm] (grow2) at (hidden2.north) {};

\draw[arrow] (rin) -- (bn.west);
\draw[arrow] (bn.east) -- (conv1.west);
\draw[arrow] (conv1.east) -- (hidden1.west);
\draw[arrow] (hidden1.east) -- (bn2.west);
\draw[arrow] (bn2.east) -- (hidden2.west);
\draw[arrow] (hidden2.east) -- (conv2.west);
\draw[arrow] (conv2.east) -- (add.west);

\draw[dashed, rounded corners, draw={rgb,255:red,160; green,110; blue,40}, fill={rgb,255:red,248; green,220; blue,180}, fill opacity=0.12]
    (-4.7,-3.8) rectangle (7.5,-0.8);
\draw[dashed, rounded corners, draw opacity=1]
    (-4.7,-3.8) rectangle (7.5,-0.8);
\node[font=\scriptsize] at (1.7,-4.0) {block 1};

\draw[thick]
    (rin) -- (-5.5,-4.5) -- (8,-4.5) -- (add.south);

\draw[dashed, rounded corners]
    (-3,-3.4) rectangle (1,-0.95);
\node[font=\scriptsize] at (-1,-3.5) {conv1};

\draw[dashed, rounded corners]
    (3,-3.4) rectangle (7,-0.95);
\node[font=\scriptsize] at (5,-3.5) {conv2};


\node[font=\scriptsize, anchor=west] (addchannels) at (0.3, 0.6) {expand hidden channels};
\draw[dashedarrow] (addchannels.south) -- (2.0, -1.3);
\draw[dashedarrow] (addchannels.south) -- (0.0, -1.3);
\draw[dashedarrow] (addchannels.south) -- (4.0, -1.3);

\node[box, basebox, minimum width=0.55cm, minimum height=0.28cm] at (5.1,-0.1) {};
\node[font=\scriptsize, anchor=west] at (5.4,-0.1) {existing channels};

\node[box, growfill, minimum width=0.55cm, minimum height=0.28cm] at (5.1,0.4) {};
\node[font=\scriptsize, anchor=west] at (5.4,0.4) {newly released channels};
\node[box, selectbox, minimum width=0.55cm, minimum height=0.28cm] at (5.1,0.9) {};
\node[font=\scriptsize, anchor=west] at (5.4,0.9) {selected block};
\node[box, scorebox, minimum width=0.55cm, minimum height=0.28cm] at (5.1,1.4) {$u_l$};
\node[font=\scriptsize, anchor=west] at (5.4,1.4) {score of the block};

\end{tikzpicture}
}
\caption{
Illustration of score-guided hidden expansion in \method{}.
The top part shows a ResNet-style CNN backbone, where adversarial spectral utilization scores $u_l$ are computed for residual blocks and the high-pressure block is selected for growth.
The bottom part expands the selected residual block: \method{} treats the hidden channels inside the residual branch as a growable parameter subspace, and the green channel blocks denote newly released capacity.
Parameter-space stabilization acts as a local stabilization mechanism in the training objective, and together with the grow mechanism forms \method{}.
}
\label{fig:pseat_overview}
\end{figure}

Figure~\ref{fig:pseat_overview} illustrates the structural component of \method{}, namely score-guided hidden expansion based on post-shortcut block representations. Parameter-space stabilization acts at the level of the training objective. Therefore, it is not drawn as a network module in the figure, but is described in the training objective below.

\subsection{Overall Framework}

\method{}, namely Guided Parameter-Space Evolution, consists of two complementary components:
\begin{enumerate}
    \item parameter-space stabilization;
    \item guided progressive hidden expansion.
\end{enumerate}

The former stabilizes local optimization within the current parameter space. The latter gradually releases new optimizable dimensions during training. In contrast to fixed-structure AT, which exposes the full parameter space from the beginning, \method{} formulates robust learning as a dynamic process of parameter-space exposure, stabilization, and expansion.

At training stage $t$, the model is determined by the current parameters $\theta_t$ and the hidden capacity configuration $d_t$:
\begin{equation}
f_t = f_{\theta_t, d_t}.
\end{equation}
Here, $d_t$ describes the hidden capacity configuration of all growable modules, and induces the current optimizable parameter space:
\begin{equation}
\Theta_t = \Theta(d_t).
\end{equation}

Thus, the training objective of \method{} can be written as:
\begin{equation}
\min_{\theta_t \in \Theta_t}
\mathbb{E}_{(x,y)\sim\data}
\left[
\loss_{\mathrm{GRAPE}}(f_{\theta_t,d_t},x,y)
\right].
\end{equation}

As training proceeds, \method{} progressively expands the optimizable parameter space through grow operations:
\begin{equation}
\Theta_0 \subset \Theta_1 \subset \cdots \subset \Theta_T.
\end{equation}

This formulation emphasizes that \method{} does not simply specify a larger model at once. Instead, it controls when the optimizable parameter space is exposed and where it is expanded during training.

\subsection{Limitation of Fixed-Parameter-Space AT}

Standard \pgdat{} optimizes within a fixed architecture and a fixed parameter space:
\begin{equation}
\theta_{t+1}
=
\theta_t
-
\eta
\nabla_\theta
\loss_{\mathrm{AT}}(\theta_t),
\qquad
\theta_t\in\Theta.
\end{equation}

where:
\begin{equation}
\Theta = \mathrm{constant}.
\end{equation}

This formulation implicitly assumes that the parameter space $\Theta$ specified before training is sufficient to support the final robust solution. It also assumes that the optimizer only needs to search for appropriate parameters within this fixed space. This paper questions whether this assumption is necessary.

In adversarial training, the model is continuously exposed to hard perturbed samples induced by the current parameters. Fixed-parameter-space training exposes all parameter degrees of freedom from the beginning. This may cause the optimizer to prematurely exploit a high-dimensional parameter space and form an inefficient robust fitting path. In contrast, if the model first learns relatively stable robust representations in a smaller parameter space and then progressively releases new optimizable dimensions, it may reach a different robust solution. Even when the final network architecture is the same, the parameter-space exposure path may affect the final training outcome.

Therefore, \method{} formulates robust training as optimization in a dynamic parameter space:
\begin{equation}
\theta_{t+1}
=
\theta_t
-
\eta
\nabla_{\theta_t}
\loss_{\mathrm{GRAPE}}(\theta_t),
\qquad
\theta_t\in\Theta_t.
\end{equation}

When a grow operation is performed, the current parameter space is expanded as:
\begin{equation}
\Theta_{t+1}
=
\Theta_t
\oplus
\Delta\Theta_{l^\ast},
\end{equation}
where $\Delta\Theta_{l^\ast}$ denotes the newly added optimizable parameter subspace in the selected module $l^\ast$.

\subsection{Parameter-Space Stabilization}

The first component of \method{} is parameter-space stabilization. It does not change the network architecture. Instead, it stabilizes adversarial training within the current parameter space $\Theta_t$. We write the training objective as:
\begin{equation}
\loss_{\mathrm{GRAPE}}
=
\loss_{\mathrm{AT}}
+
\lambda
\loss_{\mathrm{stab}},
\end{equation}
where $\loss_{\mathrm{AT}}$ is the standard PGD adversarial training loss, $\loss_{\mathrm{stab}}$ denotes the local parameter-space stabilization term, and $\lambda$ is its weighting coefficient.

In implementation, $\loss_{\mathrm{stab}}$ borrows the local smoothing idea from AWP and \iseat{}. It is used to stabilize the optimization process around highly vulnerable samples. Its goal is to reduce overly sharp loss landscapes in the current parameter space, so that the model can maintain a stable robust optimization trajectory before and after grow operations.

Specifically, $\loss_{\mathrm{stab}}$ adopts the instance-adaptive smoothing term from \iseat{}, including adaptive weighting based on instance-wise vulnerability and weight-space smoothing. All experiments use:
\begin{equation}
\lambda = 0.1,
\qquad
\gamma_{\mathrm{awp}} = 0.01.
\end{equation}

Abstractly, local stabilization can be written as:
\begin{equation}
\loss_{\mathrm{stab}}
=
\mathcal{S}
\left(
f_{\theta_t,d_t},
x,
y,
x^{adv}
\right),
\end{equation}
where $\mathcal{S}(\cdot)$ denotes a smoothing regularization term constructed from sample vulnerability or local perturbations. We do not treat this stabilization mechanism as a structural innovation. Instead, it serves as the component of \method{} that stabilizes the existing parameter space. Its complement is progressive hidden expansion, which expands new parameter space.

\subsection{Progressive Hidden Expansion}

Consider the $l$-th growable residual module:
\begin{equation}
h_{l+1}
=
S_l(h_l)
+
F_l(h_l;\theta_l,d_l),
\end{equation}
where $S_l(\cdot)$ denotes the identity shortcut or downsample shortcut, $F_l(\cdot)$ denotes the residual branch, and $d_l$ denotes the current hidden capacity inside the residual branch.

When module $l^\ast$ is selected for growth, its hidden capacity is expanded as:
\begin{equation}
d_{l^\ast}
\leftarrow
d_{l^\ast}
+
\Delta d.
\end{equation}

The corresponding residual function is expanded from:
\begin{equation}
F_{l^\ast}(h;\theta_{l^\ast},d_{l^\ast})
\end{equation}
to:
\begin{equation}
F'_{l^\ast}(h)
=
F_{l^\ast}(h;\theta_{l^\ast},d_{l^\ast})
+
G_{l^\ast}(h;\phi_{l^\ast},\Delta d),
\end{equation}
where $G_{l^\ast}$ denotes the newly added hidden subspace, and $\phi_{l^\ast}$ denotes the newly added parameters.

To avoid abruptly disrupting the function learned before growth, we use an approximately function-preserving initialization. The newly added part is initialized to produce near-zero output:
\begin{equation}
G_{l^\ast}(h;\phi_{l^\ast},\Delta d)
\approx
0.
\end{equation}

In our implementation, the newly created hidden channels are initialized conservatively. The expanded input transform of the residual branch is randomly initialized with Kaiming normal initialization, the newly added batch-normalization channels are initialized to identity statistics, and the newly added connections in the output projection are initialized to zero:
\begin{equation}
W_{\mathrm{in,new}}
\sim
\mathrm{KaimingNormal},
\qquad
(\gamma_{\mathrm{bn,new}},\beta_{\mathrm{bn,new}},\mu_{\mathrm{bn,new}},\sigma^2_{\mathrm{bn,new}})
=
(1,0,0,1),
\qquad
W_{\mathrm{out,new}}
=
0.
\end{equation}
This means that new hidden channels are created immediately, but their initial contribution to the residual output is zero because the newly added columns in the output projection are zero-connected.

Thus, the model function is approximately preserved after growth:
\begin{equation}
f_{\Theta_{t+1}}(x)
\approx
f_{\Theta_t}(x),
\end{equation}
while the optimizable parameter space has already been expanded:
\begin{equation}
\Theta_{t+1}
=
\Theta_t
\oplus
\Delta\Theta_{l^\ast}.
\end{equation}

This approximately function-preserving expansion is related to the idea of expanding networks through function-preserving transformations in Net2Net \citep{chen2016net2net}. In our case, the function-preserving effect does not rely on orthogonal initialization, but on zero-impact output coupling for the newly added hidden channels.

It is important to emphasize that progressive hidden expansion does not only change the final model capacity. Even if growth eventually restores the same architecture as fixed ResNet-18, different parameter-space exposure paths may lead to different robust solutions. Therefore, we use a sequential grow variant that preserves the final architecture as an important control. This control tests the effect of progressive parameter-space exposure itself.

\subsection{Implementation Details of Growth}

We use the eight residual blocks of ResNet-18 as growable modules. For the $l$-th block, $d_l$ denotes the hidden capacity inside the residual branch, rather than the output channel count $C_l$ after shortcut addition.

In the concrete configuration, the initial hidden capacity is set to $0.25\times$ the corresponding standard ResNet-18 block width. Each grow operation then multiplies the current hidden capacity by $4^{1/10}$. Therefore, the same block recovers the original hidden width of standard ResNet-18 after 10 grow operations.

Growth only expands the hidden subspace inside the residual branch, while keeping the external output channel count $C_l$ of the block unchanged. Thus, the dimensionality of shortcut addition remains consistent, and no modification to the shortcut branch is required. In particular, $d_l$ and $C_l$ play different roles: $C_l$ is the fixed output width of the block after shortcut addition, whereas $d_l$ is the internal hidden width of the residual branch. There is therefore no requirement that $d_l \leq C_l$; the residual branch may use either a narrower or a wider hidden subspace, as long as its output is projected back to $\mathbb{R}^{C_l\times H_l\times W_l}$. More specifically, for block $l$,
\begin{equation}
S_l(h_l),\ F_l(h_l;\theta_l,d_l),\ G_l(h_l;\phi_l,\Delta d)
\in
\mathbb{R}^{C_l\times H_l\times W_l}.
\end{equation}
Since $S_l(h_l)$ and the residual outputs before and after growth all lie in the same output space, the block output can still be written as:
\begin{equation}
h_{l+1}
=
S_l(h_l)
+
F_l(h_l;\theta_l,d_l)
+
G_l(h_l;\phi_l,\Delta d),
\end{equation}
and the dimensional compatibility of shortcut addition remains unchanged.

\subsection{Training Algorithm}

\begin{algorithm}[H]
\caption{\method{} Training}
\label{alg:grape}
\small
\begin{algorithmic}[1]
\Require Dataset $\data$; initial model $f_{\theta_0,d_0}$; total epochs $T$
\Require Grow interval $K$; growth size $\Delta d$; PGD parameters $(\epsilon,\alpha,n_{\mathrm{step}})$
\State Initialize model parameters $\theta_0$ and capacity configuration $d_0$
\For{$t = 1$ to $T$}
    \State Train $f_{\theta_t,d_t}$ for one epoch with $\loss_{\mathrm{GRAPE}}=\loss_{\mathrm{AT}}+\lambda\loss_{\mathrm{stab}}$
    \If{$t \bmod K = 0$}
        \State Generate adversarial examples using $(\epsilon,\alpha,n_{\mathrm{step}})$
        \State Extract post-shortcut outputs $Z_l^{adv}$ for all growable blocks
        \State Compute utilization scores $u_l$ for all growable blocks
        \State Select $l^\ast = \arg\max_l u_l$
        \State Expand block $l^\ast$ by $\Delta d$ with zero-impact output initialization
        \State Update parameter space $\Theta_t \rightarrow \Theta_{t+1}$
    \EndIf
\EndFor
\end{algorithmic}
\vspace{0.3em}
\noindent\textbf{Output:} Final model $f_{\theta_T,d_T}$.
\end{algorithm}

\subsection{Adversarial Spectral Utilization Score}

Progressive hidden expansion determines whether the parameter space is progressively expanded. Guided grow further determines where the parameter space should be expanded. To this end, we use an adversarial spectral utilization score to estimate the capacity pressure of different residual blocks during adversarial training.

Importantly, the score is not computed on hidden features inside the residual branch. Instead, it is computed on the output of the residual block after shortcut addition. For the $l$-th residual block:
\begin{equation}
h_{l+1}
=
S_l(h_l)
+
F_l(h_l;\theta_l,d_l),
\end{equation}
we extract adversarial representations from the block output $h_{l+1}$ to evaluate the robust representation load of this block under the current capacity configuration.

Specifically, for adversarial examples $x^{adv}$, let the output of the $l$-th residual block after shortcut addition be:
\begin{equation}
Z_l^{adv}
=
h_{l+1}(x^{adv}).
\end{equation}

After global average pooling, we obtain the batch feature matrix:
\begin{equation}
A_l^{adv}
\in
\mathbb{R}^{B\times C_l},
\end{equation}
where $B$ is the batch size and $C_l$ is the output channel count of the $l$-th block.

We center the feature matrix along the batch dimension:
\begin{equation}
\widetilde{A}_l^{adv}
=
A_l^{adv}
-
\mathbf{1}\mu_l^\top,
\qquad
\mu_l
=
\frac{1}{B}
\sum_{i=1}^{B}
A_{l,i}^{adv}.
\end{equation}

We then perform singular value decomposition on the centered feature matrix $\widetilde{A}_l^{adv}$ and obtain the singular values $\{\sigma_{l,i}\}$. These singular values are normalized into a spectral distribution:
\begin{equation}
p_{l,i}
=
\frac{\sigma_{l,i}}
{\sum_j \sigma_{l,j}+\varepsilon}.
\end{equation}

The adversarial spectral entropy is defined as:
\begin{equation}
H_l^{adv}
=
-
\sum_i
p_{l,i}
\log p_{l,i}.
\end{equation}

The effective rank is then defined as:
\begin{equation}
r_l^{adv}
=
\exp(H_l^{adv}).
\end{equation}

Effective rank is based on the spectral entropy of the singular-value distribution. It characterizes the effective dimensionality of a matrix beyond its algebraic rank \citep{roy2007effective}. In this paper, we use it to estimate the effective dimensionality of residual block output representations under adversarial examples.

Finally, we define adversarial spectral utilization as:
\begin{equation}
u_l
=
\frac{r_l^{adv}}{d_l}.
\label{eq:utilization}
\end{equation}

Here, $C_l$ is the channel count of the block output after shortcut addition, while $d_l$ is the current growable hidden capacity inside the residual branch of this block. These two quantities are not the same. The quantity $r_l^{adv}$ measures the adversarial effective dimension of the block output after shortcut addition. Therefore, $u_l$ can be interpreted as the adversarial effective dimension supported per unit hidden capacity. A higher $u_l$ indicates that the block bears a higher robust representation pressure under the current hidden capacity, and is therefore more suitable for releasing new optimizable dimensions.

At each grow event, we compute adversarial spectral utilization for all growable blocks and select the block with the highest utilization for expansion:
\begin{equation}
l^\ast
=
\arg\max_l
u_l.
\end{equation}

Accordingly, the guided grow rule of \method{} can be written as:
\begin{equation}
\Theta_{t+1}
=
\Theta_t
\oplus
\Delta \Theta_{\arg\max_l u_l}.
\end{equation}

\subsection{Parameter-Space Evolution Beyond Static Capacity}

A natural question is whether \method{} merely trains a larger model. We argue that it does not. Let the final architecture be $\mathcal{A}_{\mathrm{final}}$. Fixed-structure training optimizes in the full parameter space from the beginning:
\begin{equation}
\theta_0^{static}
\rightarrow
\theta_T^{static},
\qquad
\theta_t^{static}\in\Theta(\mathcal{A}_{\mathrm{final}}).
\end{equation}

In contrast, the training path of \method{} is:
\begin{equation}
\theta_0^{small}
\rightarrow
\theta_{t_1}^{small}
\rightarrow
\theta_{t_1}^{expanded}
\rightarrow
\cdots
\rightarrow
\theta_T^{grown}.
\end{equation}

Even if the final architecture is the same:
\begin{equation}
\Theta_T^{grown}
=
\Theta(\mathcal{A}_{\mathrm{final}}),
\end{equation}
the optimization trajectory is still different:
\begin{equation}
\theta_T^{grown}
\neq
\theta_T^{static}.
\end{equation}

Therefore, the core of \method{} is not the final parameter count itself, but the exposure and expansion path of the parameter space. The model first learns relatively stable robust representations in a smaller parameter space, and then gradually absorbs new learnable degrees of freedom. Score-guided grow further determines which high-pressure modules should receive these newly released degrees of freedom.

Moreover, since hidden expansion in different blocks leads to imbalanced growth in parameter count and FLOPs, this paper does not treat having fewer parameters as the objective by itself. We use FLOPs as the matched computation budget and examine whether guided parameter-space evolution can form a more compact and more robust parameter-space configuration under the same computation budget.

\section{Experiments}

\subsection{Experimental Setup}

We first evaluate the effectiveness of \method{} on CIFAR-10. All experiments adopt the standard $\ell_\infty$ threat model, with the perturbation budget set to:
\begin{equation}
\epsilon = 8/255.
\end{equation}

During training, we use PGD-based adversarial examples. During evaluation, we report clean accuracy and PGD-20 robust accuracy. Unless otherwise specified, all methods use the same backbone, number of training epochs, optimizer, learning rate schedule, and data augmentation settings. We use \pgdat{} on a fixed-structure ResNet-18 as the main controlled reference, and focus on comparing the robustness and parameter-space compactness of different training paths under the same or nearly matched inference FLOPs budget.

Since \method{} progressively expands hidden capacity during training, its training trajectory naturally produces intermediate models with different inference FLOPs and parameter scales. We record the intermediate models at grow stages, and use FLOPs ratio as the measure of matched inference budget. Meanwhile, we report params ratio to characterize whether the resulting parameter space is more compact under the same or nearly matched inference budget. In this paper, FLOPs ratio always refers to the inference FLOPs of the selected checkpoint relative to the fixed-structure AT reference.

For a model $m$, FLOPs ratio and params ratio are defined as:
\begin{equation}
\mathrm{FLOPsRatio}(m)
=
\frac{
F_m
}{
F_{\mathrm{AT}}
},
\end{equation}
\begin{equation}
\mathrm{ParamsRatio}(m)
=
\frac{
P_m
}{
P_{\mathrm{AT}}
},
\end{equation}
where $F_{\mathrm{AT}}$ and $P_{\mathrm{AT}}$ denote the inference FLOPs and parameter count of the fixed-structure ResNet-18 \pgdat{} reference, respectively.

The relative robust accuracy gain is defined as:
\begin{equation}
\mathrm{RobustGain}
=
\frac{
R_m - R_{\mathrm{AT}}
}{
R_{\mathrm{AT}}
}
\times 100\% ,
\end{equation}
where $R_{\mathrm{AT}}$ denotes the PGD-20 robust accuracy of the fixed-structure ResNet-18 \pgdat{} reference, and $R_m$ denotes the PGD-20 robust accuracy of the model being compared. We emphasize that the relative gain in this paper refers to the relative percentage improvement over the fixed-structure ResNet-18 \pgdat{} reference, rather than the absolute percentage-point improvement. We use inference FLOPs as the matched computation budget and report params ratio to characterize parameter-space compactness.

\begin{table}[t]
\centering
\caption{Main experimental configuration on CIFAR-10.}
\label{tab:config}
\begin{tabular}{ll}
\toprule
Item & Setting \\
\midrule
Dataset & CIFAR-10 \\
Threat model & $\ell_\infty$, $\epsilon = 8/255$ \\
Backbone & ResNet-18 \\
Evaluation attack & PGD-20 \\
Stabilization weight & $\lambda = 0.1$ \\
AWP scale in stabilization & $\gamma_{\mathrm{awp}} = 0.01$ \\
Growable modules & 8 residual blocks \\
Initial hidden ratio & $0.25\times$ of the standard block width \\
Per-grow multiplier & $4^{1/10}$ \\
Score feature position & post-shortcut block output \\
Matched-FLOPs reference & fixed-structure ResNet-18 AT \\
\bottomrule
\end{tabular}
\end{table}

\subsection{Compared Methods}

We compare the following methods:

\begin{itemize}
    \item \textbf{Fixed ResNet-18 AT.} Standard adversarial training on a fixed-structure ResNet-18, used as the main controlled reference.
    \item \textbf{AT-Grow.} A score-guided grow variant without parameter-space stabilization, used to isolate the effect of guided expansion alone.
    \item \textbf{Stabilized AT.} AT with parameter-space stabilization on a fixed-structure ResNet-18, used to evaluate the effect of local parameter-space stabilization itself.
    \item \textbf{Stabilized Seq-Grow.} A sequential grow variant with parameter-space stabilization that preserves the final ResNet-18 architecture, used to test whether progressive parameter-space exposure itself affects the final robust solution when the final architecture is the same. It grows blocks in the fixed order B1 $\rightarrow$ B2 $\rightarrow \cdots \rightarrow$ B8, rather than using score-guided selection.
    \item \textbf{\method{}.} Our proposed Guided Parameter-Space Evolution, which combines parameter-space stabilization with adversarial spectral utilization guided progressive hidden expansion.
\end{itemize}

\subsection{Main Results}

The main experiments focus on the following questions:

\begin{enumerate}
    \item At a nearly matched inference budget with a FLOPs ratio of $1.009\times$, can \method{} achieve higher PGD-20 robust accuracy than fixed-structure ResNet-18 \pgdat{}?
    \item Under matched inference FLOPs, can \method{} form a more compact parameter space with fewer parameters?
    \item Can sequential grow, which preserves the final ResNet-18 architecture, still outperform fixed-structure training, indicating that the parameter-space exposure path itself matters?
    \item Does score-guided grow further improve matched-FLOPs robust performance compared with fixed-order grow?
\end{enumerate}

\begin{table}[H]
\centering
\small
\setlength{\tabcolsep}{6pt}
\caption{Main comparison at matched FLOPs budgets on CIFAR-10.}
\label{tab:main}
\begin{tabular}{lcccc}
\toprule
Method & FLOPs Ratio & Params Ratio & Clean Acc. & PGD-20 Acc. \\
\midrule
Fixed ResNet-18 AT & 1.000$\times$ & 1.000$\times$ & 82.74 & 51.70 \\
AT-Grow & 0.996$\times$ & 0.748$\times$ & 82.82 & 53.28 \\
Stabilized AT & 1.000$\times$ & 1.000$\times$ & 81.71 & 56.05 \\
Stabilized Seq-Grow & 1.000$\times$ & 1.000$\times$ & 82.51 & 56.52 \\
\method{} & 1.009$\times$ & 0.786$\times$ & 82.43 & \textbf{56.94} \\
\bottomrule
\end{tabular}
\end{table}

The best test PGD-20 robust accuracy of the fixed-structure ResNet-18 \pgdat{} reference is 51.70\%. At a nearly matched inference budget with a FLOPs ratio of $1.009\times$, \method{} improves PGD-20 robust accuracy to 56.94\%, corresponding to a relative improvement of approximately 10.1\%, while reducing the parameter count by approximately 21.4\%. This result shows that \method{} does not simply rely on larger model capacity, but instead forms a more compact and more robust parameter-space configuration under the same inference budget.

In addition, Stabilized Seq-Grow, which preserves the final ResNet-18 architecture, reaches 56.52\% PGD-20 robust accuracy. Since this setting eventually restores the same architecture as fixed-structure ResNet-18, the improvement points to the parameter-space exposure path itself rather than only to the final model shape. Compared with Stabilized Seq-Grow, \method{} achieves slightly higher PGD-20 robust accuracy under the matched-FLOPs setting, while forming a more compact parameter configuration.

Table~\ref{tab:main} reports clean accuracy, PGD-20 robust accuracy, inference FLOPs ratio, and params ratio for different methods at key computation-budget points.

The newly completed AT-Grow result further helps isolate the effect of score-guided expansion without stabilization. Under a nearly matched inference FLOPs budget, AT-Grow reaches 53.28\% PGD-20 robust accuracy with a FLOPs ratio of $0.996\times$ and a params ratio of $0.748\times$. This indicates that guided grow alone already improves over fixed-structure AT, while the full combination of parameter-space stabilization and guided expansion in \method{} remains substantially stronger.

\subsection{Ablation Study}

We further conduct the following ablation analyses:

\paragraph{Effect of progressive parameter-space exposure.}
We compare fixed-structure ResNet-18 AT with Stabilized Seq-Grow, which preserves the final ResNet-18 architecture. Since Stabilized Seq-Grow eventually restores the same architecture as ResNet-18, higher PGD-20 robust accuracy indicates that the exposure path of the parameter space during training itself can affect the final robust solution.

\paragraph{Effect of the score-guided policy.}
We compare \method{} with Stabilized Seq-Grow to examine whether adversarial spectral utilization guided capacity release further improves matched-inference-budget robust performance compared with fixed-order capacity release under the same stabilization setting.

\paragraph{Effect of parameter-space stabilization.}
We compare fixed-structure ResNet-18 AT with Stabilized AT to evaluate the robustness gain brought by local parameter-space stabilization under a fixed-structure setting.

\paragraph{Complementarity between stabilization and guided grow.}
We compare AT-Grow with \method{} to examine whether parameter-space stabilization further improves the effectiveness of score-guided progressive hidden expansion.

\subsection{Capacity Allocation Analysis}

We further analyze the block-wise capacity allocation formed by \method{} at the selected grow stage.

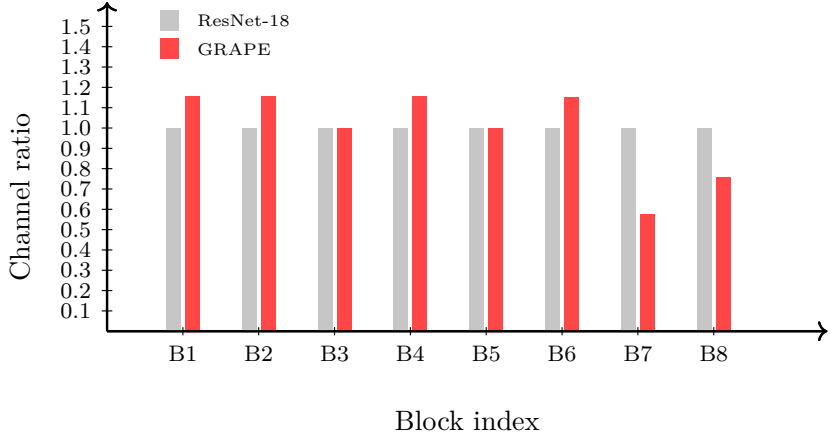
\begin{figure}[H]
\centering
\resizebox{0.68\linewidth}{!}{
\begin{tikzpicture}[x=0.95cm,y=2.55cm]
\draw[->, thick] (0,0) -- (9.5,0);
\draw[->, thick] (0,0) -- (0,1.62);
\foreach \y/\label in {0.1/0.1,0.2/0.2,0.3/0.3,0.4/0.4,0.5/0.5,0.6/0.6,0.7/0.7,0.8/0.8,0.9/0.9,1.0/1.0,1.1/1.1,1.2/1.2,1.3/1.3,1.4/1.4,1.5/1.5} {
    \draw (0.07,\y) -- (-0.07,\y) node[left] {\scriptsize \label};
}
\foreach \i/\label in {1/B1,2/B2,3/B3,4/B4,5/B5,6/B6,7/B7,8/B8} {
    \draw (\i,0.02) -- (\i,-0.02) node[below] {\scriptsize \label};
}
\node[font=\small] at (4.75,-0.44) {Block index};
\node[font=\small, rotate=90] at (-1.15,0.66) {Channel ratio};

\foreach \i in {1,...,8} {
    \fill[gray!45] (\i-0.22,0) rectangle (\i-0.03,1.0);
}
\foreach \i/\v in {1/1.1562,2/1.1562,3/1.0000,4/1.1562,5/1.0000,6/1.1523,7/0.5762,8/0.7598} {
    \fill[red!72] (\i+0.03,0) rectangle (\i+0.22,\v);
}

\draw[->, thick] (0,0) -- (9.5,0);
\draw[->, thick] (0,0) -- (0,1.62);

\fill[gray!45] (0.70,1.48) rectangle (0.95,1.58);
\node[anchor=west, font=\tiny] at (1.05,1.53) {ResNet-18};
\fill[red!72] (0.70,1.34) rectangle (0.95,1.44);
\node[anchor=west, font=\tiny] at (1.05,1.39) {\method{}};
\end{tikzpicture}
}
\caption{Block-wise channel ratios of fixed ResNet-18 and \method{} at the selected matched-FLOPs grow stage.}
\label{fig:capacity_allocation}
\end{figure}

Figure~\ref{fig:capacity_allocation} shows the block-wise channel ratios of fixed-structure ResNet-18 and \method{} under the matched-inference-FLOPs setting. We observe that \method{} does not simply recover a uniform capacity configuration, but instead forms a non-uniform hidden capacity allocation across different residual blocks.

This result indicates that, under the same or nearly matched inference budget, robust models do not necessarily require a uniform or static parameter-space configuration. Instead, through guided parameter-space evolution, the model can release more optimizable capacity to high-capacity-pressure modules, thereby forming a more compact and more robust parameter-space organization.

This analysis complements the main results: Table~\ref{tab:main} shows that \method{} improves PGD-20 robust accuracy and reduces parameter count under matched FLOPs, while Figure~\ref{fig:capacity_allocation} shows the non-uniform capacity configuration corresponding to this improvement.

\section{Discussion}

\subsection{Parameter-Space Exposure Path}

The main implication of this paper is that robust capability in adversarial training should not be explained solely by the static total number of parameters specified before training. Fixed-structure AT exposes the full parameter space from the beginning, whereas progressive grow changes when new degrees of freedom become available during optimization.

The Stabilized Seq-Grow result is the clearest evidence for this point. Stabilized Seq-Grow eventually restores the same ResNet-18 architecture as the fixed-structure AT reference, yet achieves higher PGD-20 robust accuracy. This indicates that the final architecture alone does not determine the outcome: the exposure path of the parameter space during training also matters.

From an optimization perspective, progressive grow changes how the model enters the final parameter space. Instead of searching in the full high-dimensional space from the start, the model first learns in a smaller space and then gradually absorbs new learnable capacity. This can lead to a different optimization trajectory and, ultimately, a different robust solution.

\subsection{Guided Evolution and Compactness}

\method{} adds score-guided allocation on top of progressive grow. The role of the adversarial spectral utilization score is not simply to enlarge the model, but to decide where newly released capacity should be placed.

The AT-Grow result shows that guided expansion alone is already beneficial relative to fixed-structure AT. The full \method{} result then shows that, under a nearly matched computation budget, score-guided parameter-space evolution can further improve robust accuracy while using a more compact parameter configuration. Together with the capacity-allocation figure, this suggests that compact robust models need not rely on uniform or static hidden capacity across all residual blocks.

\subsection{Relation to Robust Model Capacity}

Larger model capacity is often associated with better adversarial robustness, but our results suggest that total parameter count is only part of the story. What also matters is how capacity is released during training and which modules receive that capacity.

This perspective is consistent with prior observations that robust architectures depend not only on model size, but also on how capacity is distributed across stages \citep{huang2021architectural}. GRAPE complements that view from the training-path perspective: under matched FLOPs, it is possible to obtain a more compact parameter-space configuration without sacrificing robust performance.

\subsection{Limitations}

This work still has several limitations. First, hyperparameters such as grow interval, growth size, initialization of newly added parameters, and the definition of growable modules may affect the final results, but we have not yet systematically studied their full design space.

Second, the current experiments focus on CIFAR-10 and ResNet-18-level settings. Further evaluation on larger datasets, stronger backbones, and stricter attack protocols is still needed.

Finally, the present paper mainly provides empirical evidence that the parameter-space exposure path affects the final robust solution. A deeper theoretical analysis is needed to explain how parameter-space evolution changes adversarial optimization geometry. In addition, we mainly use PGD-20 as the controlled robustness metric here; future work should further test this framework under stronger evaluations such as AutoAttack.

\section{Conclusion}

This paper proposes \method{}, namely Guided Parameter-Space Evolution, a parameter-space evolution training method for compact adversarial robustness. Unlike fixed-structure AT, which exposes the full parameter space from the beginning of training, \method{} formulates robust model learning as a dynamic process in which the parameter space is progressively exposed, stabilized, and expanded. Specifically, \method{} combines parameter-space stabilization with progressive hidden expansion: the former stabilizes the local robust optimization process within the current parameter space, while the latter progressively releases new optimizable dimensions during training. On this basis, \method{} uses an adversarial spectral utilization score to guide newly released capacity toward high-capacity-pressure modules, thereby realizing guided parameter-space evolution.

The core conclusion of this paper is that the robust capability of adversarially trained models should not be explained solely by the static total number of parameters specified before training, but is also related to how the parameter space is exposed and expanded during training. Experiments show that, even when the final architecture is equivalent to ResNet-18, stabilized sequential grow still reaches 56.52\% PGD-20 robust accuracy, indicating that the gain does not come solely from differences in the final architecture, but is related to the parameter-space exposure path. Furthermore, at a nearly matched inference FLOPs budget with a FLOPs ratio of $1.009\times$, \method{} improves PGD-20 robust accuracy from 51.70\% for fixed-structure ResNet-18 AT to 56.94\%, corresponding to a relative improvement of approximately 10.1\%, while reducing the parameter count by approximately 21.4\%. This result suggests that guided parameter-space evolution can form a more compact and more robust parameter-space configuration under the same inference budget.

At a higher level, \method{} provides a path for adversarial training that differs from simply modifying the adversarial objective or statically enlarging model capacity. This paper does not treat parameter reduction itself as the objective. Instead, it focuses on how the parameter space can be progressively exposed, stabilized, and expanded during training under matched inference FLOPs. This perspective suggests that the effective capacity of a robust model depends not only on the final architecture, but also on the capacity release path. Future work can further evaluate the scalability of this parameter-space evolution framework on larger datasets, stronger backbones, and stricter attack evaluations, and theoretically analyze its effect on robust optimization trajectories and final robust solutions.

\section*{Acknowledgments}

This work was supported by the National Natural Science Foundation of China
(U22B2063) and the National Science and Technology Innovation 2030 Project of
China (2021ZD0202600). The model training was performed on the robotic
AI-Scientist platform of the Chinese Academy of Sciences.

\bibliographystyle{plainnat}
\bibliography{references}

\end{document}